\documentclass[letterpaper, 10 pt, conference]{ieeeconf}  

\IEEEoverridecommandlockouts                              

\usepackage{graphics} 
\usepackage{epsfig} 
\usepackage{mathptmx} 
\usepackage{times} 
\usepackage{amsmath} 
\usepackage{amssymb}  
\usepackage[mathcal]{eucal}
\usepackage{booktabs}
\usepackage{multirow}
\usepackage{makecell}
\usepackage{graphicx}
\usepackage[table]{xcolor}
\usepackage{subcaption}

\usepackage{threeparttable}

\usepackage{caption}
\usepackage{balance}

\DeclareCaptionFont{eightpt}{\fontsize{8pt}{9pt}\selectfont}

\graphicspath{ {./pic/} }
\usepackage[T1]{fontenc}

\usepackage[
backend=biber,
style=ieee,          
citestyle=numeric-comp, 
sorting=none,        
maxbibnames=3,       
minbibnames=1,
doi=false,           
isbn=false,          
url=false,           
eprint=false        
]{biblatex}

\DeclareSourcemap{
	\maps[datatype=bibtex]{
		\map{
			\step[
			fieldsource=booktitle,
			match=\regexp{\A(?:The\s+Eleventh\s+)?International\s+Conference\s+on\s+Learning\s+Representations\z},
			replace={Proc. Int. Conf. Learn. Represent. (ICLR)}
			]
		}
		\map{
			\step[
			fieldsource=booktitle,
			match=\regexp{\A(?:Proceedings\s+of\s+the\s+38th\s+)?International\s+[Cc]onference\s+on\s+[Mm]achine\s+[Ll]earning\z},
			replace={Proc. Int. Conf. Mach. Learn. (ICML)}
			]
		}
		\map{
			\step[
			fieldsource=booktitle,
			match=\regexp{\A(?:Proceedings\s+of\s+(?:[Tt]he\s+)?[0-9]+(?:st|nd|rd|th)\s+)?Conference\s+on\s+Robot\s+Learning\z},
			replace={Proc. Conf. Robot Learn. (CoRL)},
			final
			]
			\step[fieldset=organization, null]
			\step[fieldset=series, null]
		}
		\map{
			\step[
			fieldsource=booktitle,
			match=\regexp{\AReinforcement\s+Learning\s+Conference\z},
			replace={Proc. Reinforcement Learn. Conf. (RLC)}
			]
		}
		\map{
			\step[
			fieldsource=journal,
			match=\regexp{\AThe\s+International\s+Journal\s+of\s+Robotics\s+Research\z},
			replace={Int. J. Robot. Res.}
			]
		}
		\map{
			\step[
			fieldsource=journal,
			match=\regexp{\AAdvances\s+in\s+[Nn]eural\s+[Ii]nformation\s+[Pp]rocessing\s+[Ss]ystems\z},
			replace={Adv. Neural Inf. Process. Syst. (NeurIPS)}
			]
		}
		\map{
			\step[
			fieldsource=journal,
			match=\regexp{\AScience\s+Robotics\z},
			replace={Sci. Robot.}
			]
		}
		\map{
			\step[
			fieldsource=journal,
			match=\regexp{\AProceedings\s+of\s+the\s+Fourteenth\s+International\s+Conference\s+on\s+Artificial\s+Intelligence\s+and\s+Statistics\z},
			replace={Proc. Int. Conf. Artif. Intell. Stat. (AISTATS)}
			]
		}
		\map{
			\step[
			fieldsource=journal,
			match=\regexp{\AarXiv\s+preprint\s+arXiv:},
			replace={arXiv:}
			]
		}
		\map{
			\step[
			fieldsource=booktitle,
			match=\regexp{\A(?:[0-9]{4}\s+)?International\s+Conference\s+on\s+Robotics\s+and\s+Automation\s+\(ICRA\)\z},
			replace={Proc. IEEE Int. Conf. Robot. Autom. (ICRA)}
			]
		}
		\map{
			\step[
			fieldsource=booktitle,
			match=\regexp{\A(?:[0-9]{4}\s+)?IEEE\s+International\s+Symposium\s+on\s+INnovations\s+in\s+Intelligent\s+SysTems\s+and\s+Applications\s+\(INISTA\)\z},
			replace={Proc. IEEE Int. Symp. Innov. Intell. Syst. Appl. (INISTA)}
			]
		}
		\map{
			\step[
			fieldsource=journal,
			match=\regexp{\AJournal\s+of\s+[Ss]tatistical\s+Software\z},
			replace={J. Stat. Softw.}
			]
		}
		\map{
			\step[
			fieldsource=journal,
			match=\regexp{\AIEEE\s+Transactions\s+on\s+Robotics\z},
			replace={IEEE Trans. Robot.}
			]
		}
		\map{
			\step[
			fieldsource=journal,
			match=\regexp{\ACartographica:\s+the\s+international\s+journal\s+for\s+geographic\s+information\s+and\s+geovisualization\z},
			replace={Cartographica}
			]
		}
	}
}

\AtBeginDocument{
	\setlength{\abovedisplayskip}{6pt plus 2pt minus 2pt}
	\setlength{\belowdisplayskip}{6pt plus 2pt minus 2pt}
	\setlength{\abovedisplayshortskip}{3pt plus 2pt}
	\setlength{\belowdisplayshortskip}{4pt plus 2pt minus 1pt}
}

\usepackage{hyperref}

\title{\LARGE \bf
TSD: A Kinematics-Based \underline{T}rajectory \underline{S}aliency \underline{D}etector for Data Efficient Robot Imitation Learning
}

\author{Yiming Zhao$^{1}$, Gongrui Ma$^{1}$, Qingkai Li$^{1}$ and Mingguo Zhao$^{1}$
	\thanks{
			*This research was supported by STI 2030-Major Projects 2021ZD0201402 and Beijing Natural Science Foundation(L243004).%
		}
	\thanks{
			$^{1}$Department of Automation, Tsinghua University, Beijing 100084, China%
		}%
	\thanks{
			Correspondence to: Mingguo Zhao (\texttt{mgzhao@mail.tsinghua.edu.cn})
		}%
}

\begin{document}

\maketitle
\thispagestyle{empty}
\pagestyle{empty}

\begin{abstract}

Robot imitation learning relies on expert demonstrations, yet collecting complete trajectories is labor-intensive and training on every transition is computationally costly. Most pipelines allocate both budgets at the trajectory level, overlooking the distinct spatial and dynamic characteristics of task phases. We present the Trajectory Saliency Detector (TSD), a training-free method that identifies precise and agile phases directly from demonstration kinematics. TSD aligns demonstrations, detects precise operations from cross-demonstration pose dispersion, and locates agile maneuvers using centripetal acceleration. The detected phases support dataset compression by combining complete base demonstrations with salient source segments, and support dataset expansion by collecting local demonstrations at identified phases. For informative dataset compression, TSD reduces transition counts by
25.4--32.0\% and policy-training time by 13.4--26.4\%. Across five simulated tasks and two policy architectures, the resulting policies remain within 1.5 percentage points of the corresponding full-data performance or exceed it, while outperforming all budget-matched baselines. For targeted dataset expansion, TSD-guided collection outperforms matched-budget complete-trajectory collection, approaches full-data performance, and reduces collection time by 15.2\% across three real-world tasks. These results show that phase-level trajectory saliency enables more efficient allocation of demonstration and training resources. 
Project Page: \url{https://trajectory-saliency-detector.github.io/trajectory-saliency-detector/}

\end{abstract}

\section{Introduction}

Imitation learning enables robots to acquire manipulation skills from expert demonstrations, but collecting high-quality trajectories through teleoperation or kinesthetic teaching remains costly and time-consuming~\cite{real2render2real,anchordream}. Larger datasets also increase the number of transitions processed in policy training. Nevertheless, data collection and training are usually organized at the trajectory level. Operators repeatedly demonstrate a task from beginning to end, and the policy uses all recorded transitions for training without explicit phase-level prioritization. Improving data efficiency therefore calls for a better view of how human demonstration and training budgets are allocated within a trajectory.

Manipulation trajectories are internally heterogeneous~\cite{suresh2026learning, huang2026critical}. Precise phases, such as grasping and placement, require the end effector to satisfy tight spatial constraints, whereas agile phases, such as obstacle avoidance and rapid redirection, require coordinated changes in motion. Errors during these phases can directly cause failed interactions or collisions, making their coverage particularly important to task success. Transitional phases connect these operations and remain necessary for learning the complete task. However, once a base set of complete demonstrations provides the overall task context, the remaining budget can be used more effectively by concentrating additional data on salient phases with stronger precision or agility requirements, rather than repeatedly recording transitional motions.

Existing data-efficient approaches do not directly provide this phase-level allocation. Dataset-expansion methods generally increase coverage by generating or collecting additional trajectories~\cite{mimicgen,demogen,dreamgen,momagen}, but do not determine which phases should receive further demonstrations and treat different trajectory stages equally. Dataset-selection methods operate after data collection and may require learned representations or auxiliary scoring models to rank demonstrations or transitions~\cite{curating_online_exp,mi_distill,datamil,data_iwr}. Consequently, a common mechanism is needed to determine both what should be retained from an existing dataset and where additional demonstrations should be collected, directly from the shared kinematic structure of human-collected demonstrations.

\begin{figure*}[t]
	\centering
	\includegraphics[width=0.88\textwidth]{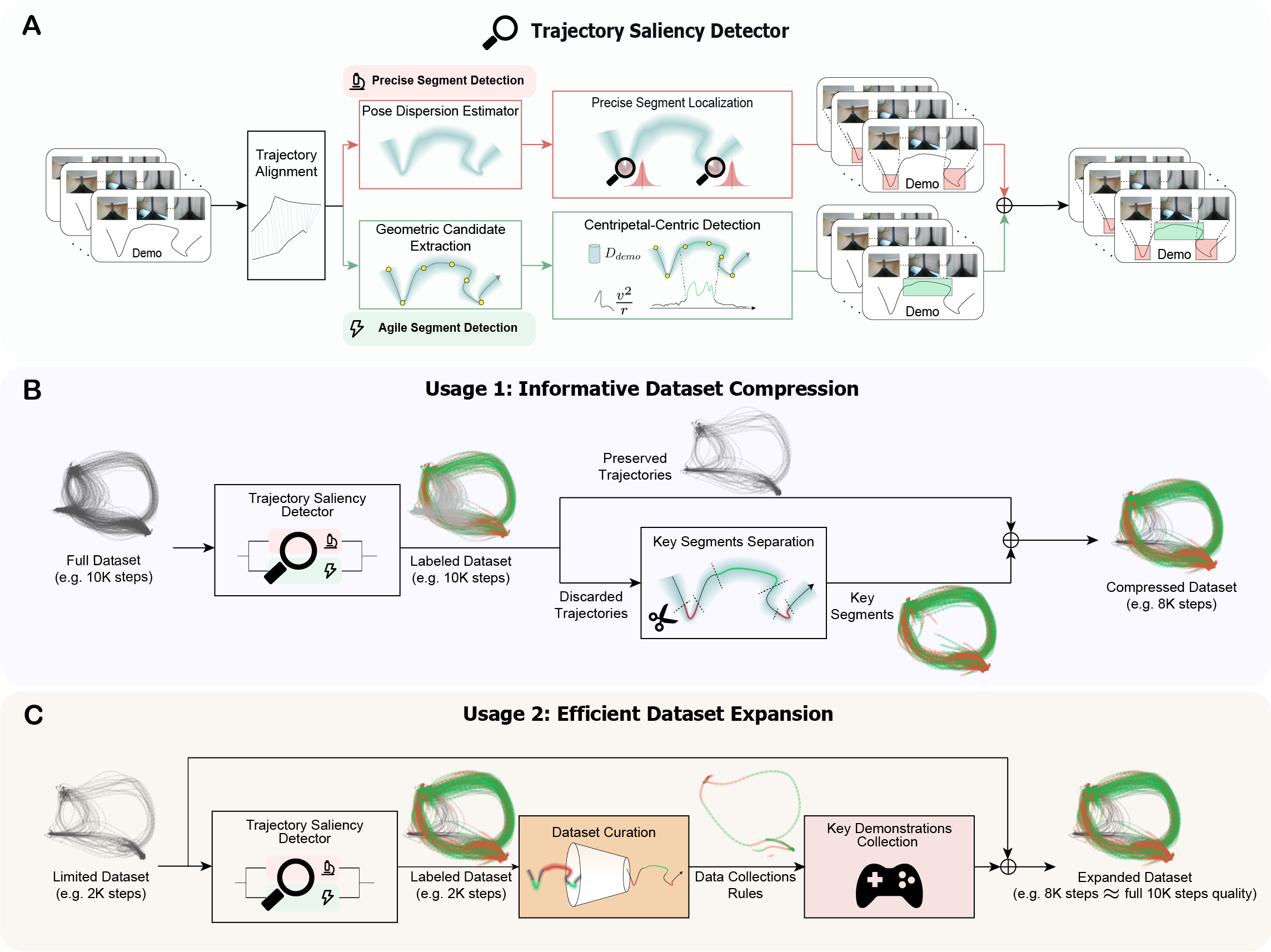}
	\caption{
		Overview of TSD. (A) TSD identifies precise segments from cross-demonstration pose dispersion and agile segments from centripetal acceleration with geometric verification. (B) For informative dataset compression, complete base demonstrations preserve task context, while only the detected segments are retained from the remaining demonstrations. (C) For efficient dataset expansion, TSD identifies the phases at which additional local demonstrations are collected and combined with the complete seed demonstrations.
	}
	\label{fig:bigpicture}
\end{figure*}

To address this problem, we propose the Trajectory Saliency Detector (TSD), a training-free method that localizes salient precise and agile phases directly from robot demonstrations, as illustrated in Fig.~\ref{fig:bigpicture}(A). TSD first aligns different demonstrations to a common task timeline. It identifies precise phases using an entropy-inspired KNN-based pose-dispersion estimator, based on the tendency of demonstrations to converge during spatially constrained operations~\cite{demospeedup}. Agile candidates are obtained from centripetal acceleration peaks and checked by geometric verification. The detected intervals are mapped back to each original trajectory and combined across detector branches and robot arms. TSD operates on trajectory kinematics and requires neither manual annotations nor auxiliary model training.

The detected phases support two complementary uses which are shown in Fig.~\ref{fig:bigpicture}(B)(C). For dataset compression, TSD retains a base set of complete demonstrations and extracts only the detected segments from the remaining source demonstrations. For dataset expansion, it uses complete seed demonstrations to identify where additional local operations should be collected. In the compression setting, TSD reduces transition counts by 25.4\% and 32.0\% for the two downstream policy across five simulated tasks in Robosuite~\cite{robosuite}, while keeping average success within 1.5 percentage points of the full-data result or exceeding it. TSD also outperforms random selection and learned data-selection baselines under comparable data scales. Across three real-world tasks, TSD-guided local collection outperforms budget-matched complete-trajectory collection, approaches the performance of full-data collection, and reduces collection time by 15.2\%. Component ablations and sensitivity analyses further examine the complementary roles of the two detector branches and the stability of the localized regions. Our main contributions are summarized as follows:
\begin{itemize}
 \item We introduce TSD, a training-free phase-level detector that localizes precise and agile phases from cross-demonstration pose dispersion and geometrically verified centripetal acceleration.

 \item We use the localized phases in a unified framework for compressing existing demonstration datasets and guiding targeted data collection, while retaining complete demonstrations for task-level context.

 \item Experiments on five simulated tasks with two policy architectures and three real-world tasks show that TSD reduces training and collection costs while preserving downstream policy performance.
\end{itemize}

\section{Related Work}
\subsection{Dataset Expansion}

Robot demonstration datasets can be expanded either by generating
additional trajectories or by collecting targeted human input.
Simulation-based methods primarily increase environmental and spatial
coverage. RoboTwin applies extensive domain randomization~\cite{robotwin2}, while rendering-based methods vary object appearance, physical properties, and materials~\cite{real2render2real,twinaligner,m3a}. Other approaches transform a small number of demonstrations across object configurations or task conditions. MimicGen reuses object-centric subtasks, DemoGen adapts action--object relations, and recent video-based methods synthesize additional observations or trajectories~\cite{mimicgen,demogen,anchordream,dreamgen,momagen}. These methods broaden the coverage of existing demonstrations, but are not designed to determine which phases of a task would benefit most from additional real-world demonstrations.

Interactive approaches reduce human effort by requesting input only during selected parts of policy execution. DAgger-style methods collect expert supervision on states visited by the learned policy, with later variants using human gating or risk to determine when intervention is needed~\cite{ross2011reduction,kelly2019hg,ThriftyDAgger}. Human-in-the-loop learning further incorporates corrective interventions or preference feedback into policy refinement and real-world reinforcement learning~\cite{luo2025precise,xia2026human,zhao2025real}. More closely related to phase-level collection, HITL-TAMP lets a task-and-motion planner execute the portions it can solve and asks the operator to demonstrate the remaining portions~\cite{mandlekar2023human}. These approaches mainly allocate human input according to the states encountered by a learned policy. In contrast, TSD identifies phases for additional collection directly from the shared kinematic patterns of seed demonstrations, before downstream policy training.

\subsection{Dataset Compression}

Demonstration selection methods reduce training data at different granularities. At the trajectory level, Demo-SCORE uses policy experience to estimate demonstration quality~\cite{curating_online_exp}, DemInf ranks demonstrations through learned state--action representations and mutual-information estimation~\cite{mi_distill}, and DataMIL estimates the contribution of individual trajectories using a pretrained datamodel~\cite{datamil}. At the segment level, retrieval methods identify task-relevant samples from prior datasets using behavior similarity, optical flow, or pretrained visual representations~\cite{behavior_retrieval,flow_retrieval,retrieval_from_prior,ram}.
IWR retrieves prior data and weights individual samples according to
their estimated relevance~\cite{data_iwr}, while L2R learns to evaluate
heterogeneous demonstration segments using preference learning~\cite{learning2discern}.

Existing methods therefore estimate data utility from policy feedback, learned representations, or task-relevance measures. TSD instead detects continuous precise and agile phases directly from the aligned kinematics of same-task demonstrations, requiring neither policy rollouts nor auxiliary representation learning.

\section{Methods}
\label{sec:method}

Let $\mathcal{D}=\{\tau_n\}_{n=1}^{N}$ denote a demonstration dataset, where the $n$-th trajectory is
\begin{equation}
	\tau_n=\left\{\mathbf{o}_{n,t},\mathbf{a}_{n,t},
	\left\{\mathbf{x}^{h}_{n,t}\right\}_{h=1}^{N_{\mathrm{arm}}}\right\}_{t=1}^{T_n}.
	\label{eq:trajectory_definition}
\end{equation}
Here, $\mathbf{o}_{n,t}$ and $\mathbf{a}_{n,t}$ are the observation and expert action, respectively, $T_n$ is the trajectory length, and $h$ indexes the robot arms. The end-effector pose is represented as $\mathbf{x}^{h}_{n,t}=[(\mathbf{p}^{h}_{n,t})^\top,(\mathbf{q}^{h}_{n,t})^\top]^\top$, where $\mathbf{p}^{h}_{n,t}\in\mathbb{R}^{3}$ is the position and $\mathbf{q}^{h}_{n,t}\in\mathbb{S}^{3}$ is a unit quaternion in $wxyz$ order. TSD uses cross-demonstration pose convergence as evidence of precise phases and geometrically verified dynamic detection as evidence of agile phases. It returns continuous intervals on the original demonstrations while leaving the downstream policy architecture and optimization objective unchanged. If demonstrations are recorded only in joint space, the required task-space poses can be obtained by forward kinematics.

\subsection{Pose-Aware Trajectory Alignment}
\label{sec:method_alignment}

Demonstrations of the same task may progress at different rates, so a common frame index does not necessarily correspond to the same task phase. We therefore use DTW\cite{dtw} to align all trajectories to a reference timeline before comparing their kinematic patterns first. We select the shortest complete demonstration as the reference, $n^\star=\operatorname*{arg\,min}_{n\in\{1,\ldots,N\}}T_n$, and set $T^\star=T_{n^\star}$, denoting the reference index and reference length, respectively. This choice yields a compact common timeline without constructing or learning a synthetic trajectory template. Robustness to the choice of reference trajectory is evaluated in Sec.~\ref{sec:reference_sensitivity}. The following sessions consider one arm and omits the arm index $h$ for simplicity. All quaternions are normalized and assigned a consistent sign.

We independently align the pose sequence $\mathbf{x}$ and the motion sequence $\boldsymbol{\xi}=[\mathbf{v}^{\top},\boldsymbol{\varpi}^{\top}]^{\top}$, where $\mathbf{v}$ and $\boldsymbol{\varpi}$ are the linear and angular velocities, respectively. We define the corresponding DTW matching costs as
\begin{align}
	d_x(\mathbf{x},\mathbf{x}')
	&=
	\lVert\mathbf{p}-\mathbf{p}'\rVert_2
	+
	2\arccos\!\left[
	\operatorname{clip}\!\left(
	\left|\mathbf{q}^{\top}\mathbf{q}'\right|,0,1
	\right)\right],
	\label{eq:pose_dtw_cost}\\
	d_{\xi}(\boldsymbol{\xi},\boldsymbol{\xi}')
	&=
	\lVert\mathbf{v}-\mathbf{v}'\rVert_2
	+
	\lVert\boldsymbol{\varpi}-\boldsymbol{\varpi}'\rVert_2.
	\label{eq:motion_dtw_cost}
\end{align}
We assign equal weights to the two terms in each cost for all tasks. Linear and angular velocities are computed using the recorded sampling interval $\Delta t$ before alignment.

Let $y\in\{x,\xi\}$ denote the alignment modality, with $\mathbf{y}$ representing the corresponding pose or motion signal. Given the $n$-th trajectory and the reference trajectory, DTW computes the cumulative alignment cost
\begin{equation}
	\begin{aligned}
		C_n^y(t,u)
		={}&
		d_y\!\left(
		\mathbf{y}_{n,t},
		\mathbf{y}_{n^\star,u}
		\right) \\
		&+
		\min\!\left\{
		\begin{array}{l}
			C_n^y(t-1,u),\quad C_n^y(t,u-1),\\[0mm]
			C_n^y(t-1,u-1)
		\end{array}
		\right\}.
	\end{aligned}
	\label{eq:dtw_recurrence}
\end{equation}
where $t$ and $u$ index frames on the original and reference timelines, respectively. We use $d_x$ for pose alignment and $d_\xi$ for motion alignment. The resulting paths are denoted by $\mathcal{W}_n^x$ and $\mathcal{W}_n^\xi$, respectively.

Both pose and motion alignment use a Sakoe--Chiba constraint to limit excessive temporal warping~\cite{dtw_window}. Each pair $(t,u)$ in a DTW path indicates that frame $t$ of the original trajectory is matched to frame $u$ of the reference trajectory. Because several original frames may be matched to the same reference frame, we collect their indices as
\begin{equation}
	\mathcal{A}_n^y(u)
	=
	\left\{
	t\mid(t,u)\in\mathcal{W}_n^y
	\right\},
	\qquad y\in\{x,\xi\}.
	\label{eq:dtw_correspondence}
\end{equation}
The matched samples are averaged to obtain the aligned pose $\widetilde{\mathbf{x}}_{n,u}$ and motion $\widetilde{\boldsymbol{\xi}}_{n,u}$. The correspondence sets are also retained to map detections from the reference timeline back to each original demonstration.

\begin{figure*}[t]
	\centering
	\includegraphics[width=0.97\textwidth]{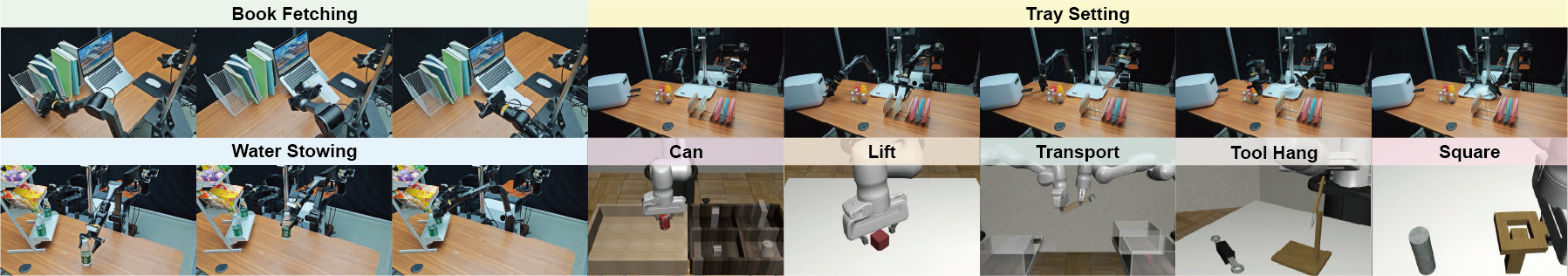}
	\caption{\textbf{Evaluation tasks.} Five robomimic simulation tasks
		and three real-world tasks cover single-arm and bimanual manipulation,
		precise object interaction, and obstacle-avoiding motion.}
	\label{fig:task_setting}
\end{figure*}
\begin{table*}[t]
	\caption{Simulation success rates (\%) for DP and ACT, evaluated over 450 randomized rollouts per condition. Pool denotes the candidate demonstration scale, and Ret. is the mean percentage of Full-100 transitions retained across the five tasks. TSD rows are shaded, and the best budget-matched average in each policy--pool setting is shown in bold.}
	\label{tab:simulation_success}
	\centering
	\footnotesize
	\renewcommand{\arraystretch}{1.02}
	\setlength{\tabcolsep}{1.8pt}
	\begin{tabular*}{\textwidth}{@{\extracolsep{\fill}}clccccccccccccc@{}}
		\toprule
		& & & \multicolumn{6}{c}{DP} & \multicolumn{6}{c}{ACT} \\
		\cmidrule(lr){4-9}\cmidrule(lr){10-15}
		Pool & Dataset & Ret. $\downarrow$ & Lift & Can & Square & Tool & Trans. & Avg. $\uparrow$ &
		Lift & Can & Square & Tool & Trans. & Avg. $\uparrow$ \\
		\midrule
		20 & Base & 20.1 & 96.6 & 53.3 & 11.1 & 5.1 & 24.6 & 38.1 &
		96.4 & 41.8 & 11.8 & 10.4 & 62.2 & 44.5 \\
		\midrule
		\multirow{6}{*}{50}
		& Random-Traj & 38.0 & 98.8 & 72.8 & 66.2 & 35.7 & 49.5 & 64.6 &
		98.7 & 64.9 & 32.7 & 25.1 & 66.9 & 57.6 \\
		& Random-Seg & 37.2 & 96.2 & 66.2 & 24.4 & 27.6 & 59.1 & 54.7 &
		99.8 & 49.8 & 26.7 & 14.0 & 46.9 & 47.4 \\
		& IWR & 39.6 & 98.7 & 79.1 & 64.2 & 18.7 & 60.2 & 64.2 &
		99.6 & 68.4 & 37.6 & 15.3 & 66.4 & 57.5 \\
		& DemInf & 37.7 & 99.1 & 81.6 & 64.7 & 20.4 & 56.4 & 64.4 &
		99.6 & 66.9 & 40.7 & 13.6 & 79.3 & 60.0 \\
		\rowcolor{gray!12}
		\cellcolor{white} & \textbf{TSD} & 37.2 & 99.3 & 82.0 & 68.8 & 35.7 & 69.5 & \textbf{71.1} &
		100.0 & 73.8 & 41.1 & 33.3 & 76.7 & \textbf{65.0} \\
		& Full-50 & 49.9 & 98.6 & 83.1 & 68.6 & 36.4 & 73.3 & 72.0 &
		99.8 & 72.0 & 38.7 & 28.7 & 70.4 & 61.9 \\
		\midrule
		\multirow{6}{*}{100}
		& Random-Traj & 68.5 & 98.6 & 91.7 & 73.3 & 46.6 & 88.8 & 79.8 &
		99.6 & 80.2 & 53.1 & 35.3 & 81.3 & 69.9 \\
		& Random-Seg & 68.7 & 98.2 & 86.2 & 38.2 & 33.3 & 73.1 & 65.8 &
		100.0 & 67.8 & 45.8 & 20.7 & 51.3 & 57.1 \\
		& IWR & 69.1 & 99.6 & 90.4 & 73.6 & 40.0 & 88.7 & 78.4 &
		100.0 & 86.0 & 52.4 & 27.1 & 72.0 & 67.5 \\
		& DemInf & 68.8 & 99.8 & 90.2 & 73.3 & 49.6 & 92.0 & 81.0 &
		99.8 & 83.1 & 56.1 & 30.9 & 78.9 & 69.8 \\
		\rowcolor{gray!12}
		\cellcolor{white} & \textbf{TSD} & 68.0 & 99.7 & 92.0 & 76.8 & 56.8 & 95.3 & \textbf{84.1} &
		100.0 & 90.7 & 61.3 & 46.9 & 81.8 & \textbf{76.1} \\
		& Full-100 & 100.0 & 99.3 & 93.7 & 77.5 & 58.6 & 89.3 & 83.7 &
		100.0 & 90.9 & 66.7 & 48.2 & 82.4 & 77.6 \\
		\bottomrule
	\end{tabular*}
\end{table*}

\subsection{Precise Segment Detection}
\label{sec:method_precise}

Based on the observed patterns that precise interactions often form spatial bottlenecks in human demonstrations~\cite{demospeedup}, we use KNN to measure the local dispersion of end-effector poses. A sliding temporal window captures how this dispersion changes throughout the task.

At each reference time $u$, we collect the aligned pose samples from all demonstrations within a window of length $L$. This produces the local sample set $\mathcal{X}_u=\{\mathbf{x}_m\}_{m=1}^{M_u}$, where $M_u=|\mathcal{X}_u|$.
We use $L=\min(L_{\max},\lfloor T^\star/10\rfloor)$ with
$L_{\max}=8$, and use the $k$-th nearest-neighbor distance with $k=10$.

For a general $d$-dimensional representation, we denote the distance to the $k$-th nearest neighbor of $\mathbf{x}_m$ by $R_k(\mathbf{x}_m)$. In our implementation, each pose contains three position and four quaternion components, and therefore $d=7$. The volume of the corresponding $d$-dimensional hypersphere is
\begin{equation}
	V(\mathbf{x}_m)=
	\frac{\pi^{d/2}}{\Gamma(d/2+1)}
	R_k(\mathbf{x}_m)^d.
	\label{eq:knn_volume}
\end{equation}
Inspired by KNN-based entropy estimation, we aggregate these local volumes into the following spatial entropy score:
\begin{equation}
	H(u)=
	\frac{1}{M_u}
	\sum_{m=1}^{M_u}
	\log\!\left(
	\frac{M_uV(\mathbf{x}_m)}{k}
	\right)
	+\psi(k)-\psi(1),
	\label{eq:pose_dispersion}
\end{equation}
where $\Gamma(\cdot)$ and $\psi(\cdot)$ denote the Gamma and digamma functions, respectively. A lower $H(u)$ indicates that the poses are more tightly concentrated across demonstrations, providing stronger evidence of a precise operation. Robustness to alternative pose metrics is evaluated in Sec.~\ref{sec:pose_metric_robustness}.

We consider continuous task demonstrations without task-irrelevant pauses, for which low end-effector speed provides additional evidence of precise operations. We therefore weight $H(u)$ using the end-effector speed. At each reference time, we first compute the local mean speed of every demonstration. Let $\bar{v}(u)$ and $\sigma_v(u)$ denote the mean and standard deviation of these values across demonstrations. We define
\begin{equation}
	\begin{aligned}
		\omega(u)
		&=
		\alpha\min\!\left\{
		1,
		\max\!\left[
		\frac{\bar{v}(u)}{v_{\max}+\epsilon},
		\lambda_v\frac{\sigma_v(u)}{\sigma_{\max}+\epsilon}
		\right]
		\right\},\\
		H_w(u)
		&=
		H_{\min}+\omega(u)\bigl[H(u)-H_{\min}\bigr],
	\end{aligned}
	\label{eq:velocity_weighted_dispersion}
\end{equation}
where $v_{\max}=\max_u\bar{v}(u)$, $\sigma_{\max}=\max_u\sigma_v(u)$, and $H_{\min}=\min_uH(u)$. The small constant $\epsilon>0$ prevents division by zero. We use $\alpha=0.9$ to control the overall weighting strength and $\lambda_v=1.0$ to balance mean speed and cross-demonstration speed variation. Consequently, regions that are both slow and speed-consistent receive lower scores.

We next convert the weighted dispersion curve into precise-operation candidates. We detect valleys of $H_w(u)$ by applying peak detection to $f(u)=-H_w(u)$. To adapt the detector to signals and trajectories of different scales, the minimum peak prominence is set to $\beta[\max_u f(u)-\min_u f(u)]$, while the minimum peak distance and width are set to $\max(1,\lfloor\rho_D T^\star\rfloor)$ and $\max(1,\lfloor\rho_W T^\star\rfloor)$ to suppress repeated or short-lived detection. Prominence is measured relative to the dispersion curve, allowing TSD to remain identifiable
when the world-frame locations of task-objects vary across demonstrations. The accepted peak locations define the precise centers on the reference timeline $n^\star$. The same scale-adaptive rule is later applied to the acceleration signal for agile detection. The coefficients $\beta$, $\rho_D$, and $\rho_W$ are fixed across tasks and shared by both detectors.

\begin{table*}[t]
	\centering
	\begin{minipage}[t]{0.685\textwidth}
		\centering
		\captionof{table}{Real-world dataset expansion with Diffusion Policy. Success rates are evaluated over 30 trials per condition; size is the percentage of the task-specific Full-100 transitions. TSD columns are shaded, and the best budget-matched success rate in each task and pool is shown in bold.}
		\label{tab:real_world_results}
		\footnotesize
		\setlength{\tabcolsep}{1.5pt}
		\renewcommand{\arraystretch}{1.01}
		\begin{tabular*}{\linewidth}{@{\extracolsep{\fill}}llccccccc@{}}
			\toprule
			& & & \multicolumn{3}{c}{50\% pool} & \multicolumn{3}{c}{100\% pool} \\
			\cmidrule(lr){4-6}\cmidrule(lr){7-9}
			Task & Metric & Base & Rand.-Traj & \cellcolor{gray!12}\textbf{TSD} & Full & Rand.-Traj & \cellcolor{gray!12}\textbf{TSD} & Full \\
			\midrule
			\multirow{2}{*}{Tray Setting}
			& Suc. (\%) $\uparrow$ & 0.0 & 46.7 & \cellcolor{gray!12}\textbf{60.0} & 53.3 & 76.7 & \cellcolor{gray!12}\textbf{86.6} & 83.3 \\
			& Size (\%) $\downarrow$ & 21.0 & 45.3 & \cellcolor{gray!12}45.0 & 50.6 & 84.9 & \cellcolor{gray!12}84.9 & 100.0 \\
			\midrule
			\multirow{2}{*}{Water Stowing}
			& Suc. (\%) $\uparrow$ & 40.0 & 53.3 & \cellcolor{gray!12}\textbf{56.7} & 63.3 & \textbf{70.0} & \cellcolor{gray!12}\textbf{70.0} & 76.6 \\
			& Size (\%) $\downarrow$ & 20.5 & 40.4 & \cellcolor{gray!12}40.8 & 50.3 & 75.6 & \cellcolor{gray!12}75.8 & 100.0 \\
			\midrule
			\multirow{2}{*}{Book Fetching}
			& Suc. (\%) $\uparrow$ & 46.7 & 53.3 & \cellcolor{gray!12}\textbf{60.0} & 66.7 & 76.7 & \cellcolor{gray!12}\textbf{93.3} & 93.3 \\
			& Size (\%) $\downarrow$ & 19.7 & 42.5 & \cellcolor{gray!12}42.3 & 49.7 & 80.1 & \cellcolor{gray!12}80.3 & 100.0 \\
			\bottomrule
		\end{tabular*}
	\end{minipage}
	\hfill
	\begin{minipage}[t]{0.295\textwidth}
		\centering
		\captionof{table}{Additional real-world data collection time. Complete and TSD-guided data were collected separately. Times include phase resetting, teleoperation, and exclude collection of the shared base set.}
		\label{tab:real_collection_time}
		\footnotesize
		\setlength{\tabcolsep}{1.5pt}
		\renewcommand{\arraystretch}{1.02}
		\begin{tabular*}{\linewidth}{@{\extracolsep{\fill}}lccc@{}}
			\toprule
			Task & Complete & TSD & Save \\
			& (min) & (min) & (\%) \\
			\midrule
			Tray Setting & 52.5 & 46.6 & 11.2 \\
			Water Stowing & 43.5 & 35.0 & 19.5 \\
			Book Fetching & 54.0 & 46.0 & 14.8 \\
			\midrule
			Mean & 50.0 & 42.5 & 15.2 \\
			\bottomrule
		\end{tabular*}
	\end{minipage}
\end{table*}

\subsection{Agile Segment Detection}
\label{sec:method_agile}

We define an agile motion as a finite trajectory interval in which the end effector rapidly redirects its translational motion~\cite{agility}. For a path $\mathbf{p}(t)$ with speed $v(t)$ and curvature $\kappa(t)$, the centripetal-acceleration magnitude is $a_c(t)=v(t)^2\kappa(t)$. This quantity remains small for both fast straight motion and slowly traversed bends. It becomes large when a pronounced turn is executed rapidly, making it suitable as the primary signal for detecting agile motion.

For the $n$-th demonstration, let $\mathbf{v}_{n,t}$ denote the linear velocity. Its time derivative $\dot{\mathbf{v}}_{n,t}$ is computed using the recorded sampling interval. We calculate the centripetal-acceleration magnitude as
\begin{equation}
	a_{c,n}(t)=
	\begin{cases}
		\dfrac{\left\|\mathbf{v}_{n,t}\times
			\dot{\mathbf{v}}_{n,t}\right\|_2}
		{\left\|\mathbf{v}_{n,t}\right\|_2},
		& \left\|\mathbf{v}_{n,t}\right\|_2>\epsilon_v,\\[8pt]
		0, & \text{otherwise},
	\end{cases}
	\label{eq:centripetal_acceleration}
\end{equation}
where $\epsilon_v>0$ avoids numerical instability. The signal is mapped to the reference timeline using DTW correspondence $\mathcal{A}_n^\xi$. We apply the shared rule described in \ref{sec:method_precise} to aligned $\widetilde{a}_{c,n}(u)$. The temporal support of each accepted peak is extracted to produce the candidate interval set $\mathcal{I}_{n}^{A}$.

Because $a_c$ is computed from temporal derivatives, small frame-to-frame fluctuations in demonstration data can be amplified and produce spurious peaks. We therefore apply geometric verification to the position trajectory before accepting an acceleration interval. Specifically, RDP~\cite{rdp} simplifies the pose-aligned three-dimensional path and retains points at which it deviates substantially from a straight-line approximation. We further evaluate the tortuosity of a position interval $[s,e]$:
\begin{equation}
	\zeta_n(s,e)=
	\left(\textstyle\sum_{u=s}^{e-1}
	\left\|\widetilde{\mathbf p}_{n,u+1}
	-\widetilde{\mathbf p}_{n,u}\right\|_2\right)
	\Big/
	\left\|\widetilde{\mathbf p}_{n,e}
	-\widetilde{\mathbf p}_{n,s}\right\|_2.
	\label{eq:tortuosity}
\end{equation}
This ratio equals one for a straight path and increases as the path becomes more circuitous. Intervals with negligible displacement are excluded from the tortuosity test to avoid numerical instability. We compute the tortuosity of intervals defined by the RDP points. Each high-tortuosity interval votes for its endpoints, and an RDP point is retained if it is supported by multiple such intervals. This removes isolated geometric deviations while preserving points that consistently delimit curved motion.
The remaining points form the geometrically verified set $\mathcal{Q}_{n}^{f}$.

Finally, an acceleration interval in $\mathcal{I}_{n}^{A}$ is retained only if it contains at least one point in $\mathcal{Q}_{n}^{f}$. The retained intervals are the detected agile segments on the reference timeline. Because dynamic and geometric evidence is evaluated along each demonstration, agile phases can be identified across different spatial routes representing multimodal behavior.

\subsection{Segment Integration and Dataset Construction}
\label{sec:method_integration}

The detected precise and agile regions are initially expressed on the reference timeline. We map them back to each original demonstration using their corresponding DTW alignment. For $y\in\{x,\xi\}$, the inverse mapping is
\begin{equation}
	\phi_n^y(u)=\operatorname{median}\!\left(
	\bigcup_{\ell=\max(1,u-\delta_m)}^{\min(T^\star,u+\delta_m)}
	\mathcal{A}_n^y(\ell)\right),
	\label{eq:median_mapping}
\end{equation}
where $\delta_m$ defines a small neighborhood around reference index $u$. The median reduces the influence of one-to-many DTW matches. A precise center $c$ is mapped by $\phi_n^x(c)$, whereas an agile interval $[s,e]$ is mapped by its motion-alignment endpoints $[\phi_n^\xi(s),\phi_n^\xi(e)]$.

Because the precise detector identifies the center of an operation, we extend each mapped center in both time and space. The temporal and axis-wise spatial tolerance are
\begin{equation}
	\begin{aligned}
		g_n &=
		\operatorname{clip}\!\left(
		\left\lfloor\rho_gT_n\right\rfloor,
		g_{\min},g_{\max}\right),\\
		r_{n,j} &=
		\rho_s\left(
		\max_t p_{n,t,j}-\min_t p_{n,t,j}
		\right),\qquad j\in\{1,2,3\}.
	\end{aligned}
	\label{eq:precise_extension}
\end{equation}
A temporal window of length $g_n$ is placed around the mapped center. We also extend the segment while the end effector remains within $r_{n,j}$ of the center along every position axis. The final precise segment covers both extensions, retaining local context and adapting its duration to each demonstration. Agile intervals already have boundaries determined by the acceleration peak and therefore require only endpoint mapping.

Overlapping or adjacent precise and agile intervals from all robot arms are merged. For dataset compression, the resulting segments from the source demonstrations are combined with the complete base demonstrations. For dataset expansion, they specify the task phases for which additional local demonstrations are collected.

\begin{figure*}[t]
	\centering
	\begin{subfigure}[t]{0.64\textwidth}
		\centering
		\includegraphics[width=\linewidth]{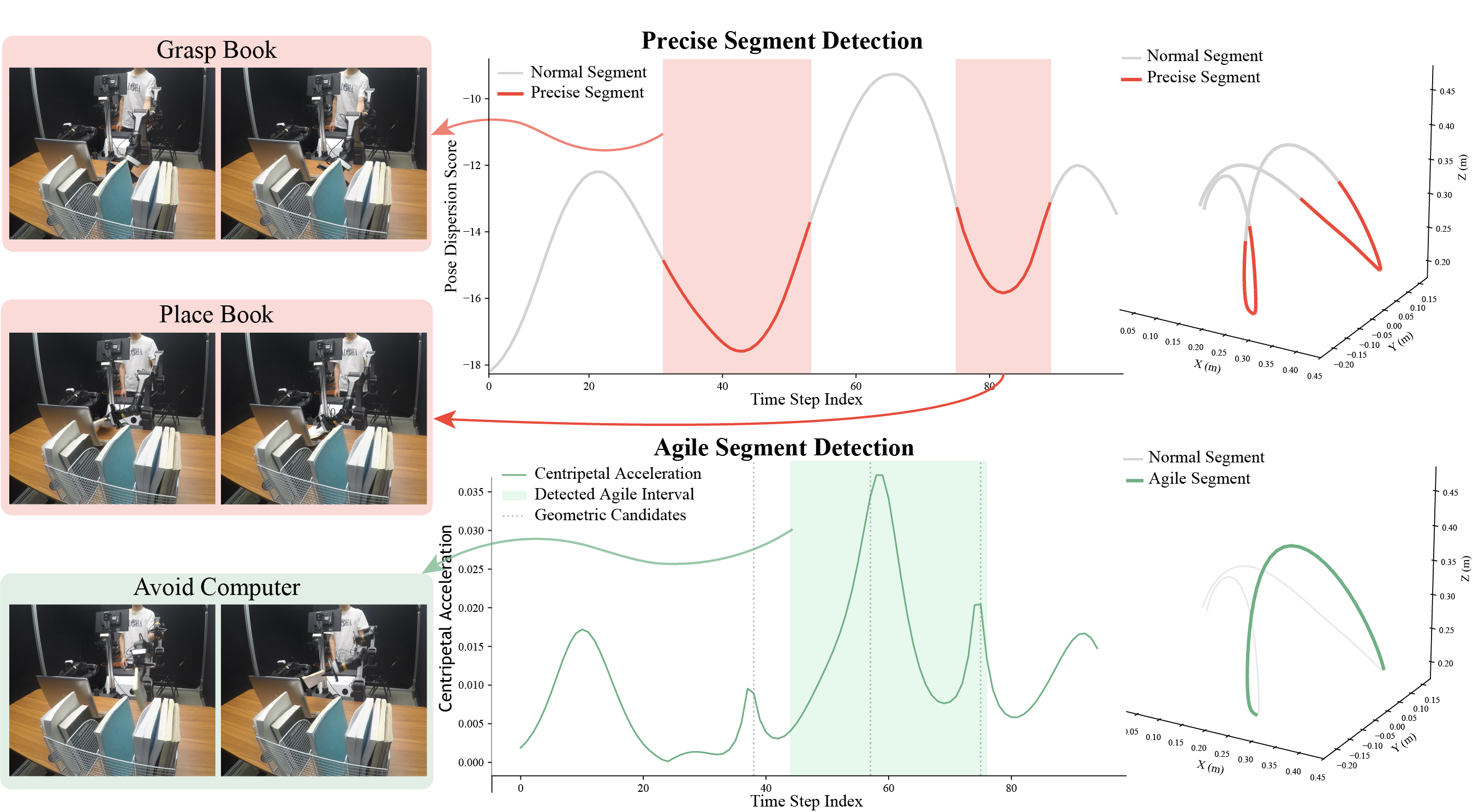}
		\caption{TSD localization on Book Fetching.}
		\label{fig:qualitative}
	\end{subfigure}
	\hfill
	\begin{subfigure}[t]{0.33\textwidth}
		\centering
		\includegraphics[width=\linewidth]{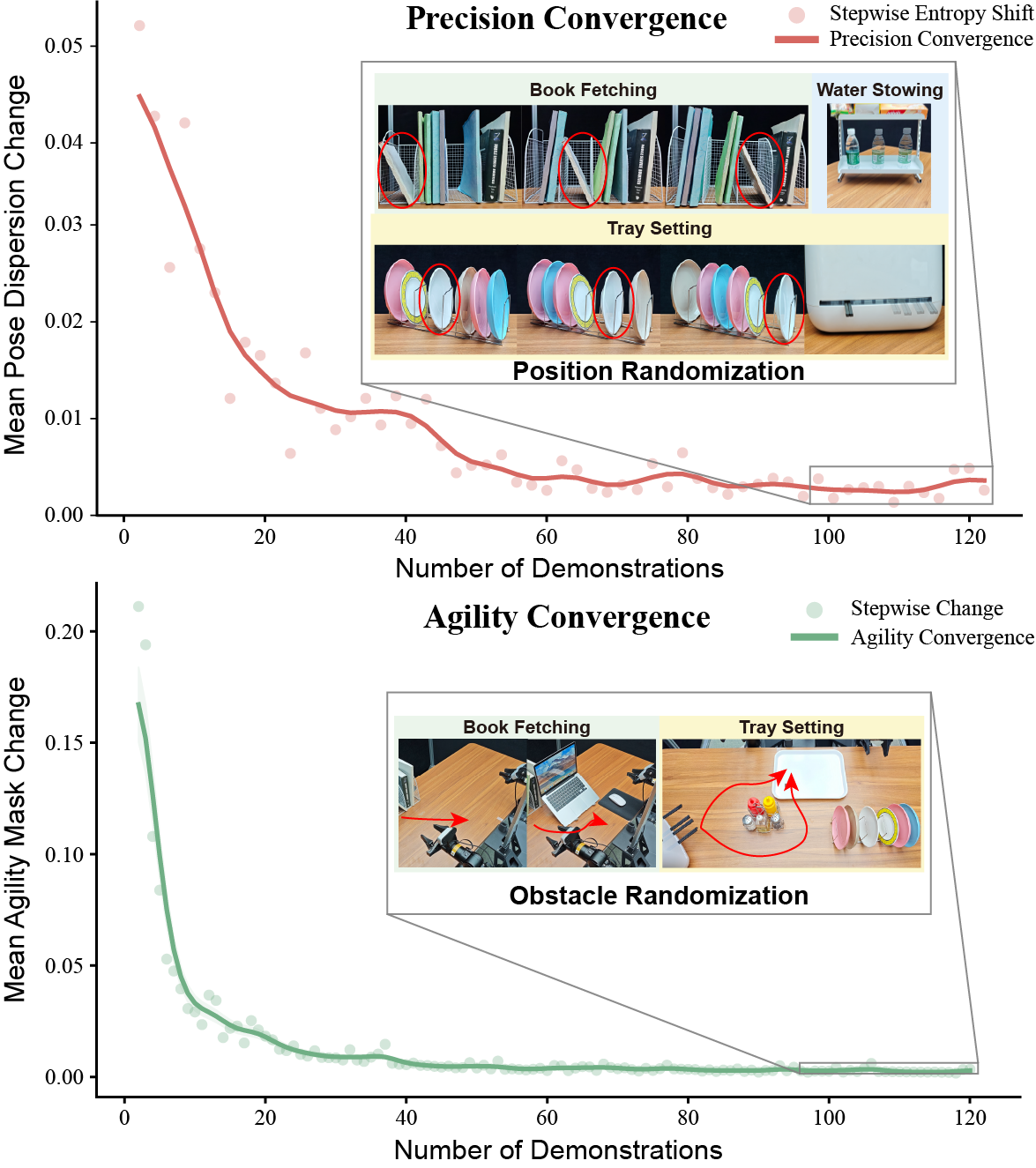}
		\caption{Detection convergence.}
		\label{fig:convergence}
	\end{subfigure}
	\caption{\textbf{Qualitative localization and convergence.} 
		(a) Precise (red) and agile (green) intervals detected in Book Fetching, with corresponding observations and end-effector paths.
		(b) Mean absolute changes in the precise dispersion curve and agile detection mask as demonstrations are added; lower values indicate convergence.
		The insets show randomized task-object positions and alternative left/right obstacle-avoidance routes; the detections in (a) are obtained from demonstrations collected under these variations.}
\end{figure*}

\section{Experiments}
\label{sec:experiments}

\subsection{Experimental Setup}
\label{sec:experimental_setup}

\textbf{Tasks and evaluation protocol.}
We evaluate TSD for \textbf{dataset compression} on five robomimic tasks including Lift, Can, Square, Tool Hang, and Transport~\cite{robomimic} using robosuite task randomization~\cite{robosuite}. \textbf{Dataset expansion} is evaluated on three real-world tasks. Book Fetching and Water Stowing combine single-arm obstacle avoidance with precise grasping or placement, while Tray Setting requires two arms to retrieve and place a plate and chopsticks around condiment bottles. Figures~\ref{fig:task_setting} and~\ref{fig:convergence} show the tasks settings and real-world randomization.

\textbf{Dataset construction and baselines.}
Each simulation task contains 200 complete demonstrations. We use pools of 100 (50\%) and 200 (100\%) demonstrations to test whether compression depends on source data scale. The same 40 complete demonstrations form the base set, and the remaining 60 or 160 form the source set. TSD combines the base with detected source segments to produce \emph{TSD-50} and \emph{TSD-100}. All baselines use the same partition: \emph{Full} retains the complete pool, \emph{Random-Traj} selects complete source trajectories close to TSD's transition count, \emph{Random-Seg} replaces each TSD segment with a
non-overlapping random segment of the same length from the same trajectory. IWR~\cite{data_iwr} selects timesteps using estimated importance, and DemInf~\cite{mi_distill} ranks complete demonstrations using learned state--action representations.

Real-world tasks use the same two scales, with 20\% complete demonstrations as the base set. Segment and complete-trajectory data are collected separately. Because pick-and-place tasks have reproducible intermediate robot--object configurations, the operator can reset each detected phase and demonstrate only the corresponding local operation. Baseline instead collect complete trajectories. \emph{Random-Traj} selects trajectories up to TSD's transition count, whereas \emph{Full-50} and \emph{Full-100} retain the corresponding complete pools.

\textbf{Method Settings.}
Simulation experiments use both image-based Diffusion Policy(DP)~\cite{dp} and Action Chunking with Transformers(ACT)~\cite{act}; real-world experiments use DP. The visual input contains a third-person view and wrist-mounted views; real-world images have resolution $240\times320$. All policies are trained for 1,200 epochs on an NVIDIA GeForce RTX 5090 GPU, with each epoch traversing the corresponding training set. For each simulation condition, the success rate is computed from 450 rollouts across 3 seeds. For each real-world condition, the success rate is computed from 30 trials. TSD uses a unified detector configuration for all simulation and real-world tasks, without task-specific tuning. The three shared peak-detection coefficients are $\beta=0.10$, $\rho_D=0.05$, and $\rho_W=0.02$.

\begin{figure}[t]
	\centering
	\begin{subfigure}{0.48\columnwidth}
		\centering
		\includegraphics[width=\linewidth]{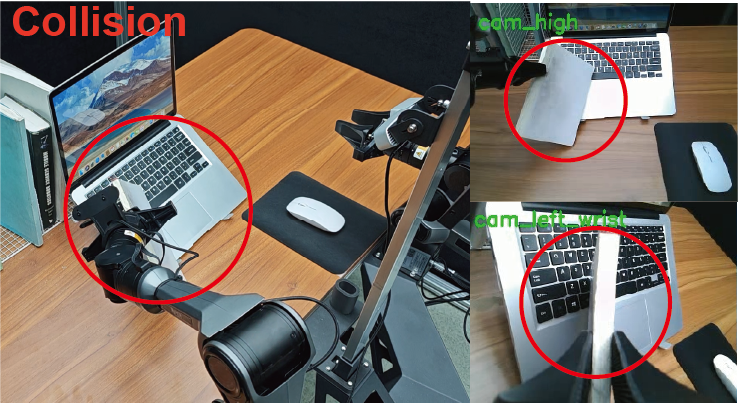}
		\caption{Without agile segments}
		\label{fig:only_precision}
	\end{subfigure}
	\hfill
	\begin{subfigure}{0.48\columnwidth}
		\centering
		\includegraphics[width=\linewidth]{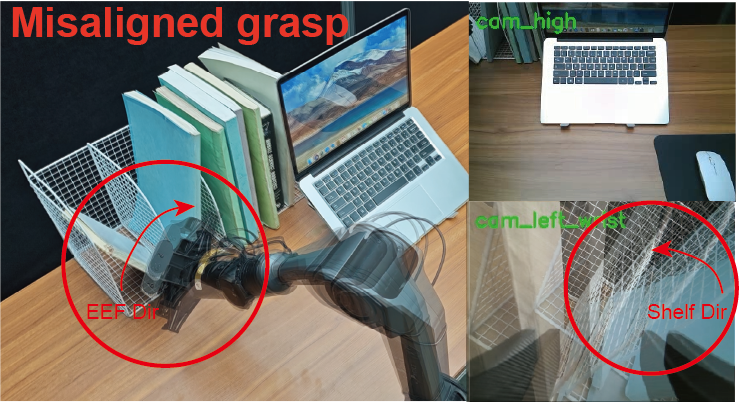}
		\caption{Without precise segments}
		\label{fig:only_agility}
	\end{subfigure}
	\caption{\textbf{Representative single-branch failures on Book Fetching.} Removing agile segments leads to hesitation or collision during obstacle avoidance, whereas removing precise segments leads to inaccurate grasping.}
	\label{fig:ablation}
\end{figure}
\begin{table}[t]
	\caption{Ablation study on two representative tasks. Success
		rates are percentages; P and A denote precise and agile segments; ``matched'' denotes complete-trajectory
		controls with similar transition counts.}
	\label{tab:ablation}
	\centering
	\footnotesize
	\setlength{\tabcolsep}{3pt}
	\renewcommand{\arraystretch}{1.01}
	\begin{tabular*}{\columnwidth}{@{\extracolsep{\fill}}lcccc@{}}
		\toprule
		& \multicolumn{2}{c}{Book Fetching} &
		\multicolumn{2}{c}{Tool Hang} \\
		\cmidrule(lr){2-3}\cmidrule(lr){4-5}
		Dataset & Transitions & Succ.(\%) & Transitions & Succ.(\%) \\
		\midrule
		\textbf{TSD (P+A)} & 14921 & 93.3 & 64237 & 56.8 \\
		\midrule
		Precise only & 12938 & 63.3 & 62283 & 52.9 \\
		Full-Traj (P-matched) & 13015 & 70.0 & 60930 & 46.2 \\
		\midrule
		Agile only & 7220 & 40.0 & 22179 & 16.0 \\
		Full-Traj (A-matched) & 7110 & 50.0 & 22870 & 30.0 \\
		\bottomrule
	\end{tabular*}
\end{table}

\begin{table*}[t]
	\centering
	\begin{minipage}[t]{0.315\textwidth}
		\centering
		\captionof{table}{Pose-metric robustness averaged over simulation tasks. $\Delta$Ret. denotes the change in the retained-data ratio in percentage points.}
		\label{tab:detector_robustness}
		\footnotesize
		\setlength{\tabcolsep}{2.0pt}
		\renewcommand{\arraystretch}{1.01}
		\begin{tabular*}{\linewidth}{@{\extracolsep{\fill}}lccc@{}}
			\toprule
			Metric & IoU & F1 & $\Delta$Ret.(\%) \\
			\midrule
			7D Euclidean & 1.000 & 1.000 & 0.000 \\
			SE(3), $\eta{=}0.5$ & 0.925 & 0.961 & 1.458 \\
			SE(3), $\eta{=}1$ & 0.949 & 0.981 & 0.806 \\
			SE(3), $\eta{=}2$ & 0.994 & 0.997 & $-0.048$ \\
			Quat. chordal & 0.949 & 0.981 & 0.806 \\
			\bottomrule
		\end{tabular*}
	\end{minipage}
	\hfill
	\begin{minipage}[t]{0.315\textwidth}
		\centering
		\captionof{table}{Reference sensitivity over 10 alternative reference per task. A hit denotes a normalized Chamfer distance below 5\%.}
		\label{tab:reference_sensitivity}
		\footnotesize
		\setlength{\tabcolsep}{2.0pt}
		\renewcommand{\arraystretch}{1.01}
		\begin{tabular*}{\linewidth}{@{\extracolsep{\fill}}lcc@{}}
			\toprule
			Task & Hit rate & Norm. Chamfer \\
			\midrule
			Lift & $0.9780\pm0.0207$ & $0.0091\pm0.0031$ \\
			Can & $0.9730\pm0.0131$ & $0.0102\pm0.0033$ \\
			Square & $0.9340\pm0.0255$ & $0.0162\pm0.0038$ \\
			Tool Hang & $0.9082\pm0.0363$ & $0.0150\pm0.0059$ \\
			Transport & $0.9101\pm0.0475$ & $0.0170\pm0.0036$ \\
			\bottomrule
		\end{tabular*}
	\end{minipage}
	\hfill
	\begin{minipage}[t]{0.315\textwidth}
		\centering
		\captionof{table}{Parameter sensitivity measured by mean frame IoU over simulation tasks.}
		\label{tab:parameter_sensitivity}
		\footnotesize
		\setlength{\tabcolsep}{1.8pt}
		\renewcommand{\arraystretch}{1.01}
		\begin{tabular*}{\linewidth}{@{\extracolsep{\fill}}lccc@{}}
			\toprule
			Setting & Prec. & Agile & Union \\
			\midrule
			$\beta=0.05$ & $1.000$ & $0.993$ & $1.000$ \\
			$\beta=0.20$ & $0.841$ & $0.972$ & $0.843$ \\
			$L_{\max}=4$ & $0.934$ & -- & $0.941$ \\
			$L_{\max}=16$ & $0.930$ & -- & $0.935$ \\
			$(\rho_D,\rho_W){=}(0.025,0.010)$ & $0.998$ & $0.973$ & $0.997$ \\
			$(\rho_D,\rho_W){=}(0.100,0.040)$ & $0.982$ & $0.715$ & $0.975$ \\
			\bottomrule
		\end{tabular*}
	\end{minipage}
\end{table*}
\begin{table}[t]
	\caption{Computational efficiency. Policy training time is averaged
		over tasks.  Save is the mean$\pm$ standard deviation of the task-wise training-time reduction, and Perf. is the TSD-to-Full success-rate ratio.}
	\label{tab:efficiency}
	\centering
	\footnotesize
	\setlength{\tabcolsep}{3pt}
	\renewcommand{\arraystretch}{1.01}
	\textbf{(a) Downstream policy training}\\[1mm]
	\begin{tabular*}{\columnwidth}{@{\extracolsep{\fill}}lccccc@{}}
		\toprule
		Setting & Pool & Full (h) & TSD (h) & Save (\%) & Perf. (\%) \\
		\midrule
		Sim.--DP & 50 & 9.43 & 7.77 & $13.4\pm4.5$ & 98.7 \\
		Sim.--DP & 100 & 17.99 & 13.46 & $19.9\pm5.4$ & 100.5 \\
		Sim.--ACT & 50 & 13.03 & 10.83 & $19.7\pm8.8$ & 105.0 \\
		Sim.--ACT & 100 & 26.54 & 20.29 & $26.4\pm11.2$ & 98.1 \\
		\midrule
		Real--DP & 50 & 15.86 & 13.45 & $14.5\pm2.2$ & 96.4 \\
		Real--DP & 100 & 28.15 & 23.67 & $16.5\pm3.6$ & 98.7 \\
		\bottomrule
	\end{tabular*}\\[1.5mm]
	\textbf{(b) Dataset curation time}\\[1mm]
	\begin{tabular*}{\columnwidth}{@{\extracolsep{\fill}}llcc@{}}
		\toprule
		Method & Computation & Mean time & Range \\
		\midrule
		TSD & Trajectory processing & $42.51\pm49.76$ s & 2.02--106.80 s \\
		DemInf & Two VAE stages & 1.54 h & 1.46--1.74 h \\
		IWR & State--action VAE & 2.56 h & 2.55--2.59 h \\
		\bottomrule
	\end{tabular*}
\end{table}

\subsection{Simulation: Informative Dataset Compression}
\label{sec:simulation_experiments}

Table~\ref{tab:simulation_success} shows the compression benefit of TSD persists across the two candidate-pool scales. TSD-50 and TSD-100 use 25.4\% and 32.0\% fewer transitions than their Full-50 and Full-100 counterparts respectively. This size reduction is not accompanied by a decrease in policy performance. Relative to Full-50 and Full-100, TSD changes the five-task average success rate by $-0.9$ and $+0.4$ percentage points with DP, and by $+3.1$ and $-1.5$ points with ACT. Thus, in all four settings, TSD either exceeds the corresponding complete dataset or remains within 1.5 percentage points of it, with substantially fewer transitions.

Budget-matched baselines comparisons evaluate the effect of data selection. TSD exceeds Random-Traj by 4.3--7.3 percentage points, showing that concentrating the data budget on salient phases is more effective than retaining fewer complete trajectories. Random-Seg trails Random-Traj by 9.9--14.0 points and TSD by 16.4--19.0 points, showing that segmentation alone is insufficient. The selected phases determine the benefit. TSD also outperforms IWR and DemInf in every policy--pool setting without requiring an auxiliary scoring model.
These comparisons show that the compression benefit comes from identifying which parts of the demonstrations should be retained, rather than simply reducing the dataset or representing it as a collection of segments.

\subsection{Real-World: TSD-Guided Dataset Expansion}
\label{sec:real_world_experiments}

The real-world study tests whether TSD can guide additional data collection more effectively than complete collection. We compare TSD with budget matched Random-Traj and use Full-50 and Full-100 as full data references using DP. IWR and DemInf cannot guide new collection because they require pre-collected data. Random-Seg is omitted because it is consistently weaker than other methods in simulation.
Figures~\ref{fig:qualitative} and~\ref{fig:convergence} provide an example from the randomization used across the real-world tasks. Despite randomized task-object positions and alternative routes, TSD still localizes grasping and placement as precise phases and obstacle avoidance as an agile phase. Thus, TSD tolerates varying workspace poses and geometric paths. These intervals guide the operator to collect local demonstration phases.

Using the phases detected in this manner, Table~\ref{tab:real_world_results} shows that TSD raises mean success from 28.9\% with the base set to 58.9\% and 83.3\% at the 50\% and 100\% scales. It exceeds Random-Traj by 7.8 and 8.8 percentage points and remains within 2.2 and 1.1 points of Full-50 and Full-100 respectively, while using only 40.8--45.0\% and 75.8--84.9\% of the Full-100 transitions. Thus, directing new demonstrations to detected phases provides greater benefit than collecting a similar budget of complete trajectories and can approach full-data performance with fewer transitions.

Table~\ref{tab:real_collection_time} further shows that TSD guided local collection reduces average demonstration time from 50.0 to 42.5 min. This 15.2\% reduction shows that the transition savings translate into less operator demonstration time, demonstrating that TSD detected phases provide effective guidance for data collection.

\subsection{Detector Analysis}
\label{sec:detector_analysis}

\textbf{Component ablation.}
Table~\ref{tab:ablation} examines the contribution of each detector branch. Complete TSD achieves the highest success on both tasks. Precise-only data remain close to TSD on Tool Hang but cause a 30.0 point drop on Book Fetching, while agile-only data perform substantially worse on both tasks. Thus, the precise branch provides most of the standalone performance, but neither branch
alone reproduces their combination.

The matched controls and Fig.~\ref{fig:ablation} clarify their different roles. Precise-only data outperform the matched control on Tool Hang and comparable on Book Fetching, whereas agile-only data trail the size matched policies on both tasks. Qualitatively, removing precise segments causes grasp drift, while removing agile segments causes hesitation or collision during obstacle avoidance. These results show that TSD benefits from retaining both precise interactions and agile redirections instead of prioritizing either type alone.

\textbf{Detection convergence.}
We examine the stability of TSD detection as more demonstrations are added under task-object and obstacle randomization. We measure consecutive changes in the pose-dispersion curve and the aligned binary agile mask. As shown in Fig.~\ref{fig:convergence}, the agile output reaches a low plateau after approximately 20 demonstrations, while the precise dispersion stabilizes more gradually. The diminishing changes show that both branches converge as the dataset grows despite varying interaction locations and avoidance routes.

\textbf{Pose-metric robustness.}
\label{sec:pose_metric_robustness}
To examine whether precise-phase localization depends on the raw-7D Euclidean distance used by KNN, we replace this distance while keeping all other detector settings fixed. The alternative $\mathrm{SE}(3)$-aware metric is
\begin{equation}
	d_{\eta}^{G}(\mathbf{x},\mathbf{x}')
	=
	\sqrt{
		\left(
		\frac{\lVert\mathbf{p}-\mathbf{p}'\rVert_2}{s_p}
		\right)^2
		+
		\eta^2
		\left(
		\frac{2\arccos\!\left(|\mathbf{q}^{\top}\mathbf{q}'|\right)}{s_R}
		\right)^2},
	\label{eq:se3_knn_metric}
\end{equation}
We obtain $s_p$ and $s_R$ by sampling pairs of end-effector poses from the demonstrations and taking the median position and rotation differences, respectively. We evaluate $\eta=0.5$, $1$, and $2$, corresponding to reduced, nominal, and increased rotational weighting, respectively. We also replace rotation with the sign-invariant quaternion chordal distance $\sqrt{2-2|\mathbf{q}^{\top}\mathbf{q}'|}$.

Across the alternative metrics, Table~\ref{tab:detector_robustness} reports center F1 scores of 0.961--0.997, frame IoUs of 0.925--0.994, and changes in retained data of at most 1.46 percentage points. The normalized geodesic and chordal metrics produce identical intervals. These results show that precise-phase localization is not specific to single distance metric.

\textbf{Reference sensitivity.}
\label{sec:reference_sensitivity}
Since TSD aligns demonstrations to a selected reference, we examine whether the reference choice affects the localized salient regions. Using the shortest demonstration as the default, we rerun TSD with 10 alternative references per task while keeping all other settings fixed. Different references produce different timelines, so their detected segments cannot be compared by frame index. Instead, we compare their end-effector paths using symmetric Chamfer distance. The distance is normalized by the trajectory workspace diagonal. Table~\ref{tab:reference_sensitivity} shows spatial match rates above 90\% and mean normalized distances below 1.71\% for all five tasks, indicating that TSD identifies similar workspace regions under different reference choices.

\textbf{Parameter sensitivity.}
Table~\ref{tab:parameter_sensitivity} compares each variant with the default detection using mean frame IoU. Lowering $\beta$ or halving the temporal constraints leaves the union mask nearly unchanged, while varying $L_{\max}$ retains union IoUs above 0.93. Sensitivity appears mainly under stricter settings. $\beta=0.20$ suppresses weak precise valleys, and doubled temporal constraints remove short agile peaks. Thus, TSD is robust to relaxed thresholds and changes in the entropy window, whereas overly restrictive settings produce more conservative detections as expected.

\subsection{Computational Efficiency}
\label{sec:computational_efficiency}

Table~\ref{tab:efficiency} tests whether the reduction in transitions produces practical computational savings. TSD reduces simulation training time by 13.4--26.4\% while retaining 98.1--105.0\% of full-data performance. In the real-world experiments, it reduces training time by 14.5--16.5\% with 96.4--98.7\% performance retention. The computation overhead does not offset these savings. TSD processes a simulation dataset in $42.51\pm49.76$ s on average, whereas DemInf and IWR require 1.54 h and 2.56 h, respectively, to train their auxiliary representations. TSD therefore provides a computationally lightweight way to reduce the cost of downstream policy training.

\section{CONCLUSIONS}

We presented the Trajectory Saliency Detector (TSD), a training-free method that localizes precise and agile phases from aligned demonstration kinematics and uses them for dataset compression and targeted data collection. Under comparable data budgets, TSD outperforms alternative selection and collection strategies, while achieving comparable success rate to policies trained on the corresponding complete datasets. Future work will extend phase-level saliency detection to dexterous and loco-manipulation tasks.






\balance
\printbibliography[title={References}]

\end{document}